\documentclass[journal]{IEEEtran}
\usepackage{graphicx}
\usepackage{cite}
\usepackage{picinpar}
\usepackage{amsfonts}
\usepackage{amsmath}
\usepackage{url}
\usepackage{flushend}
\usepackage[latin1]{inputenc}
\usepackage{colortbl}
\usepackage{soul}
\usepackage{multirow}
\usepackage{pifont}
\usepackage{color}
\usepackage{alltt}
\usepackage[hidelinks]{hyperref}
\usepackage{enumerate}
\usepackage{siunitx}
\usepackage{breakurl}
\usepackage{epstopdf}
\usepackage{pbox}
\usepackage{subcaption}
\usepackage[labelformat=simple]{subcaption}

\begin{document}
\title{Associated Spatio-Temporal Capsule Network for Gait Recognition}


\author{
	\vskip 1em
	{
	Aite Zhao,
	Junyu Dong,
	Jianbo Li,
	Lin Qi,
	and Huiyu Zhou
	}

	\thanks{
		{
		Aite Zhao and Jianbo Li are with College of Computer Science and Technology, Qingdao University, China. E-mail: zhaoaite@qdu.edu.cn; lijianbo@qdu.edu.cn.
		
		Junyu Dong and Lin Qi are with College of Information Science and Engineering, Ocean University of China, China. E-mail:dongjunyu@ouc.edu.cn; qilin@ouc.edu.cn.
		
		Huiyu Zhou is with School of Informatics, University of Leicester, United Kingdom. E-mail: hz143@leicester.ac.uk.
		
        (Corresponding author: Jianbo Li.)
		}
	}
}

\maketitle

\begin{abstract}
It is a challenging task to identify a person based on her/his gait patterns. State-of-the-art approaches rely on the analysis of temporal or spatial characteristics of gait, and gait recognition is usually performed on single modality data (such as images, skeleton joint coordinates, or force signals). Evidence has shown that using multi-modality data is more conducive to gait research. Therefore, we here establish an automated learning system, with an associated spatio-temporal capsule network (ASTCapsNet) trained on multi-sensor datasets, to analyze multimodal information for gait recognition. Specifically, we first design a low-level feature extractor and a high-level feature extractor for spatio-temporal feature extraction of gait with a novel recurrent memory unit and a relationship layer. Subsequently, a Bayesian model is employed for the decision-making of class labels. Extensive experiments on several public datasets (normal and abnormal gait) validate the effectiveness of the proposed ASTCapsNet, compared against several state-of-the-art methods.
\end{abstract}

\begin{IEEEkeywords}
Gait recognition, associated capsules, spatio-temporal, capsule network, multi-sensor.
\end{IEEEkeywords}

{}

\definecolor{limegreen}{rgb}{0.2, 0.8, 0.2}
\definecolor{forestgreen}{rgb}{0.13, 0.55, 0.13}
\definecolor{greenhtml}{rgb}{0.0, 0.5, 0.0}

\section{Introduction}

\IEEEPARstart{N}{owadays}, there are a number of technologies for identifying a person, and one of them is face recognition, which has been applied in a wide range of applications \cite{Luan2017Disentangled,Jingna2020An}. Additionally, fingerprint recognition, palmprint and action recognition have attracted large attention from the research community \cite{Cao2017Automated,Jia2017Palmprint,LiuDWDYL20}. Gait recognition is a biometric technology that recognizes people at a longer distance by acquiring walking data \cite{Bregler1997Learning,Minxiang2020Distinct}, which aims to analyze the overall human activity rather than part of the human body, and gait is not easily imitated and hence more reliable for the recognition task.



Human beings have a unique visual system, which can directly identify people based on human gait beyond a certain distance. The input data of a gait recognition model is generally a sequence of walking videos or images, and its data collection is similar to face recognition, which is noninvasive. However, due to a large amount of data from image sequences, the computational complexity of gait recognition is relatively high, and the processing is not straightforward. A gait recognition system extracts key features from the images of the joint movement of a walking person. So far, there is no commercialized gait-based identity authentication system.

In previous research, many types of sensors have been used for gait data collection and recognition - image, depth and inertial sensors, which acquire RGB images/videos, motion distance and acceleration/orientation. Among them, captured images have the advantages of convenience, non-contact, and easy understanding. Besides, skeleton data well preserve the integrity of gait movement whilst minimizing the interference of color and shape of the human body in the image. Some existing work can directly extract 3D skeleton nodes from the images of the human body and this character makes up for the shortcomings of 2D RGB images \cite{Liu2017Skeleton,Li2019Attentive}.


On the one hand, traditional gait recognition algorithms have been well studied including Hough transform, contour extraction, and three-dimensional wavelet moments. Model-based algorithms are well investigated in multiple data sources. Besides, deep learning methods, such as convolutional neural networks (CNN) and restricted Boltzmann machine (RBM) have been employed and achieved promising results in gait recognition or other relevant fields \cite{Wu2016A,Hossain2013Multimodal,Qi2016Hierarchically}. Moreover, there are several studies on image segmentation and image classification, which can be applied to the field of gait recognition to solve the problem of multimodal gait data classification \cite{2018Image,2018An}. However, these approaches are not designed to deal with sequential data.


On the other hand, although some feature extraction algorithms take temporal changes into account, they lack the analysis of spatial information. These methods use the relationship between time series to represent the dynamics in the gait cycle, but do not analyze each frame in detail. Specifically, there are several state-of-the-art methods for sequential information acquisition \cite{hochreiter1997long,10.5555/3305381.3305543,Shu,8397466,Zhao2018Dual,Liu2017Skeleton}, such as original, bi-directional, and attention-based LSTM. However, these methods come from the standard LSTM in order to enhance its performance, but the spatial mapping of the input data has not been fully addressed.

Apart from temporal and spatial characteristics, we also consider the structural relationship between the various parts of the human body when analyzing gait, such as the relationship between feet and legs, head and body, and limbs and trunk. Unlike CNN, capsule network (CapsNet) \cite{Sabour2017Dynamic} does not bypass the position of the entity (body parts) within the region (human body). For low-level capsules (body parts), location information is ``place-coded" whereas capsule is active. Using an iterative routing process, each active capsule will choose a capsule in the layer to be its parent in the tree. For the task of gait recognition, this iterative process will be solving the problem of locating the body parts.

In order to extract temporal, spatial, and structural features from multimodal data, we design a capsule network to synchronously analyze gait patterns and changes. It mainly includes two tasks, feature extraction and classification. We extract and classify features, and evaluate the effectiveness of feature extraction in classification. Inspired by the learning process of CapsNet \cite{Sabour2017Dynamic} and the gated recurrent unit (GRU) \cite{hochreiter1997long}, an associated spatio-temporal capsule network (ASTCapsNet) is here proposed for gait recognition, including three modules: a low-level feature extractor, a high-level feature extractor, and a decision-making layer.

The low-level feature extractor aims to extract spatio-temporal features of the input data, including a novel expandable memory module and a convolution layer; the high-level feature extractor mainly performs matrix operation on the feature map in the form of capsules, including the primary capsule layer, the relationship layer and the digital capsule layer with the dynamic routing algorithm.

In the decision-making layer, we design four softmax classifiers to analyze the output of all layers, and the parameters of each layer of the network are optimised until the redefined joint loss converges to the minimum. Moreover, the standard Bayesian model is employed to fit the results of the classifiers and regain as a new feature input. Furthermore, we also propose a relationship layer between the capsules to calculate the relative relationship between the local features of gait.

The main contributions of this paper are summarized as follows:

\begin{itemize}
	\item An associated spatio-temporal capsule network (ASTCapsNet) is proposed for spatio-temporal and structural feature extraction and classification using multimodal gait data.
	\item Two feature extractors are designed for low- and high-level gait feature analysis and extraction. The low-level feature extractor contains a novel memory module and a convolution layer to extract spatio-temporal features. In the high-level feature extractor, we propose a relationship layer to calculate the relationship matrix between two capsules to ensure the structural integrity of the input.
	\item We develop a recurrent memory unit to enable the model to converge faster than the original GRU unit.
	\item The decision-making layer is used to construct the evaluation mechanism of ASTCapsNet, through four softmax classifiers and a Bayesian model to select optimal features.
    \item Compared against the other state-of-the-art technologies, the proposed method achieves superior results on five challenging datasets, including both normal and abnormal gait data, for multimodal gait recognition.
\end{itemize}

This paper is organized as follows. Section II describes existing studies on gait recognition. Section III introduces the proposed ASTCapsNet model, and Section IV shows experimental results. We discuss the performance of ASTCapsNet, and point out its weaknesses in Section V. We conclude the whole paper in Section VI.


\section{Related Work}
Significant efforts have been made to progress the state-of-the-art of gait recognition. From handcrafted feature recognition to model-based recognition, gait recognition methods have undergone significant progress. It must be pointed out that gait data collection process has become diverse.

\subsection{Multi-Sensor Based Methods}
There are several ways to obtain gait data collected by sensors. The types and output of the sensors are different and can be roughly classified into the following types: cameras, infrared, depth, force sensors, accelerators and gyroscopes.

Camera acquisition is convenient and non-invasive, which can capture people's gait form at a certain distance. Chattopadhyay \textit{et al.} combined front and back view features from RGB-D images. The data acquisition was conducted at the airport security checkpoint with two depth cameras installed on top of the metal detector door outside the yellow line \cite{Chattopadhyay2014Frontal}. Xue \textit{et al.} used an infrared thermal imaging camera to collect gait images to establish an infrared thermal gait database, which can detect the human body and remove noise from complex background \cite{Xue2010Infrared}.


There is also the use of gyroscopes and accelerators to develop wearable measurement modules for measuring inertial signals generated while walking and to assess gait, as a basis for new hemiplegia diagnostic techniques using wearable devices \cite{Park2017Design}.

Besides, some studies were based on foot force datasets. Force sensors were installed on the sole to capture the foot movement data of a patient in order to monitor the progress of neurodegenerative diseases \cite{Zhao2018Dual}.

\subsection{Hand-crafted Features Based Approaches}
Based on the above-mentioned data sources, early gait recognition techniques often use traditional computer vision methods for image understanding and produce hand-crafted features from raw data. Gianaria \textit{et al.}\cite{gianaria2014human} defined a dataset of physical and behavioral features to identify relevant parameters for gait description. Andersson \textit{et al.} \cite{andersson2015person} acquired gait attributes in each gait cycle, such as leg angle, step length, stride length, and anthropometric attributes such as mean and standard deviation.

However, these features are in a low-dimensional space and are mainly based on human experience. Therefore, it is difficult to obtain descriptive features in a high-dimensional space.

\subsection{Model-Based Approaches}
With remarkable advantages that deep learning plays in various fields, gait recognition technologies have also employed deep learning algorithms to model the dynamic changes of gait. When observing human walking, people can describe the structure of the human body and detect motion patterns of human limbs. The structure of the human body can be interpreted based on their prior knowledge. Model-based gait recognition methods play an active role in recent development. 

Feng \textit{et al.} proposed a new feature learning method to learn temporal information in a gait sequence for cross-view gait recognition. Heat maps extracted by a CNN-based pose estimation method were used to describe the gait information in every frame, and the standard LSTM was adopted to model a gait sequence \cite{Feng2017Learning}.

Wu \textit{et al.} utilized deep CNNs for human identification via similarity learning, which can recognize the most discriminative changes of gait patterns \cite{Wu2016A}. In order to extract gait spatio-temporal features, CNN and LSTM have also been simultaneously applied to establishing models for gait recognition \cite{Alotaibi2017Improved}. Although these methods explain gait patterns well, the characteristics of gait are not only the stacking of features, but also include structural and positional information, such as the positional relationship between arms and legs, and so on.

The activities of the neurons within CapsNet represent the various properties of a particular entity that is present in the data. These properties can include many different types of instantiation parameters such as pose, deformation, velocity, and texture, which can improve gait recognition. Even though Capsule Network has been applied \cite{Zhao_2019_CVPR,NIPS2018_7823}, gait recognition is seldomly involved.


In this paper, we propose an associated spatio-temporal capsule network (ASTCapsNet) to process the input gait
data and learn more powerful multi-level structured representations at each class of the streaming gait data. The newly proposed low- and high-level feature extractors enhance the capability of our framework for gait recognition in different gait streams. Moreover, we introduce softmax classifiers and the Bayesian model to evaluate the importance of the features generated by each module, which increases the accuracy of classification. Furthermore, we extensively evaluate the proposed gait recognition framework over five datasets, including the KinectUNITO, SDUgait, CASIA-A datasets for people identification, and NDDs, PD datasets for disease diagnosis.

\section{Associated Spatio-Temporal Capsule Network}

\begin{figure*}[htp]
   \centering
   \includegraphics[width=13cm]{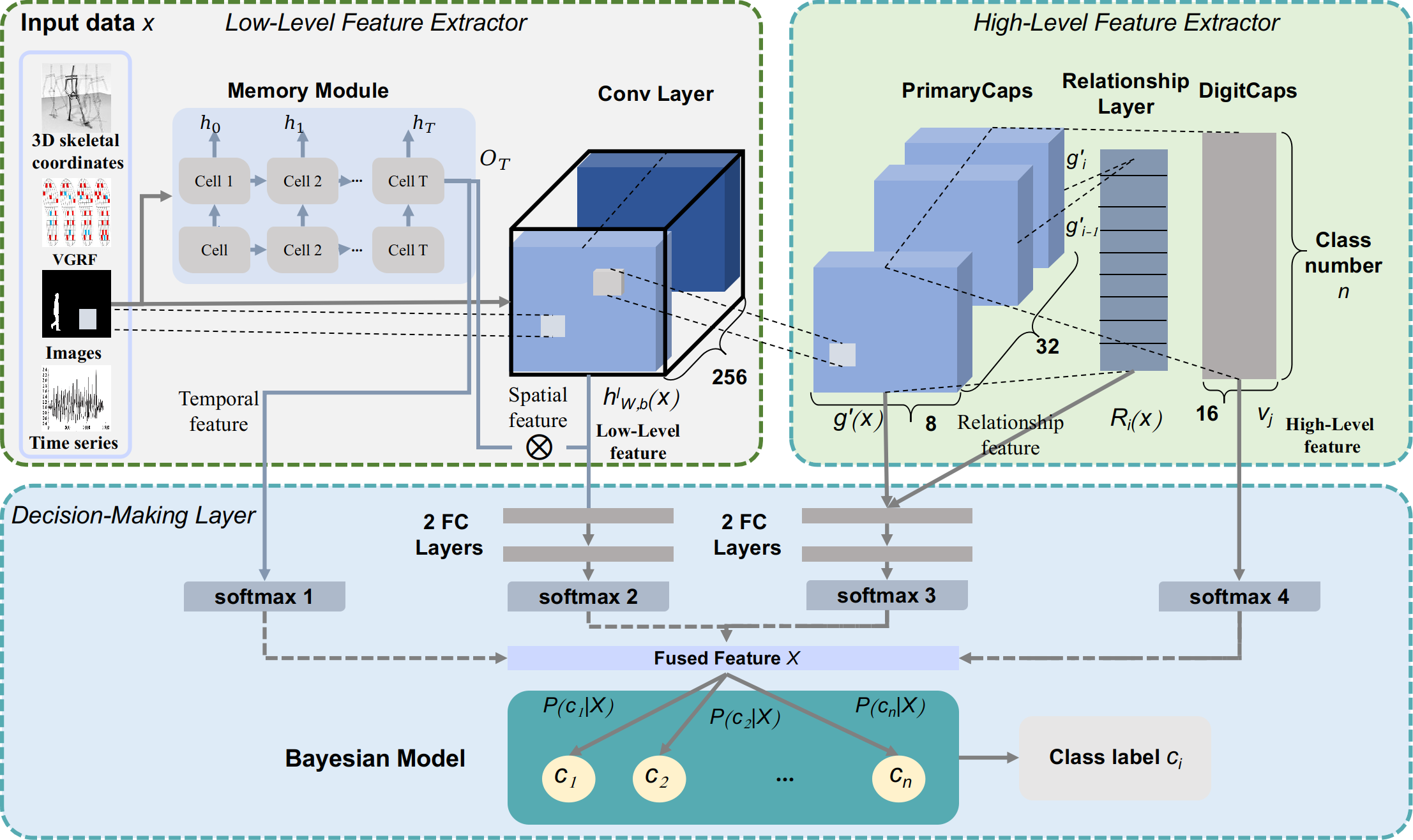}
   \caption{The structure of ASTCapsNet. It is divided into three modules: A low-level feature extractor for spatio-temporal feature modeling, including a memory module and a convolution layer; a high-level feature extractor for learning structural and relationship features, including a primary capsule layer, a relationship layer, and a digit capsule layer; and a decision-making layer for classification, including four softmax classifiers and a Bayesian model. Four softmax classifiers behind the first two modules are utilized to evaluate the system performance. The Bayesian model utilizes the outputs of the previous two feature extractors to determine the final class labels. }
   \label{capsnet_process}
\end{figure*}

We introduce the proposed architecture, associated spatio-temporal capsule network (ASTCapsNet), for gait recognition in this section. The overall schema of this method is illustrated in Fig. \ref{capsnet_process}. Here, we take the input gait images as an example to illustrate that other sensor data also follows this step. Firstly, the image is fed into the memory module and the convolution layer at the same time, and the generated spatio-temporal features are used as the input of softmax 1 and 2. Then, after the high-level feature extractor receives the low-level spatio-temporal features, the capsule layer and the relationship layer are used to extract the high-level structure and relationship features as the input of softmax 3 and 4, and the decision-making module utilizes the fused feature for the final decision.

In the proposed network, the inputs of ASTCapsNet are continuous sensor output data at each time step including images, force-sensitive data, 3d skeleton joint data, and time series. The structure of the gated recurrent unit (GRU) is performed in the time-domain to model the gait dynamics over data frames.  To effectively extract high-level features from low-level features and determine the relationship between them, a dynamic routing algorithm is used for weighing operations. To better deal with different levels of feature extracted by the network, a Bayesian modeling mechanism is also introduced to process the output of each layer for our network.

On the one hand, for extracting spatial information from gait, CNN (convolutional neural network) is one of the most popular methods which can implement feature extraction by accumulating adjustments layer by layer. However, in this process, there is an important information loss - the relative position and relationship between features \cite{Sabour2017Dynamic}. Precisely it is since the pooling operation in CNN can only provide rough position information, allowing the model to ignore small spatial changes, and cannot accurately learn the position correlation of different objects, and to this end, the capsule network (CapsNet) emerged.



On the other hand, for capturing the temporal information, gated recurrent unit (GRU) is a recurrent neural network that can be expanded into several cells (a chainlike model), which can extract information at each time step, and the connection between hidden units is iterative.


\subsection{Low-Level Feature Extractor}
Capsule network (CapsNet) \cite{Sabour2017Dynamic} has verified their superior strength in extracting a spatial feature from different data types \cite{Zhao_2019_CVPR}.  Inspired by the success of these approaches in the analysis of different data, we use a convolutional mechanism and capsule structure in CapsNet to extract spatial features. Besides, by adopting the advantages of processing temporal sequential data of GRU, we design a temporal mask added in each layer to represent the unique spatial and temporal features. Specifically, a hierarchical architecture with a spatio-temporal feature extraction layer is utilized in our model to learn a comprehensive representation.

To introduce the ASTCapsNet recurrent neural networks, we begin with the definition of several notations. Suppose we are given an input sequence $D = {\{x_i \in \mathbb{R}^N, i=1,2,...,M\}}$, where $M$ is the sample number and $N$ is the feature dimension in each sample, with corresponding class label sequences $L={\{y_i \in \mathbb{R}^1, i=1,2,...,M\}}$. For the task of gait recognition, our aim is to address three problems: (1) Successfully obtaining spatio-temporal feature. (2) Acquiring high-level features, and (3) seeking the final class label.

\subsubsection{Convolution Layer}
The first building block of our network model is the convolution layer. The purpose of this layer is to use several sliding convolution kernels with a step size of 1 to filter the input data and extract multiple feature maps.

Firstly, we utilize $L_2$ normalization to scale the original input $D=\{x_i\in\mathbb{R}^N, i=1,2,...,M\}$ to [0,1] with the output of $D'$. And then, we feed $D'=\{x'_i\in\mathbb{R}^N, i=1,2,...,M\}$ into memory and convolution modules, respectively. For the pre-processing of $D'$, we reshape each training sample in a format of $K \times T = N$, where $K$ and $T$ are the numbers of rows and columns of each sample so that convolution operation can be carried out in the form of matrix no matter what the data type is. The sizes of the convolutional kernels must be smaller than $K$ and $T$ respectively. The feature map is demonstrated as follows:
\begin{equation}
{h^{l}_{W,b}(x')} =f(W^Tx')= f\left(\sum_{i=1}^M{W_i}^{l-1}*{x'_i}^{(l-1)}+b^{l-1}\right).
\label{conv_1}
\end{equation}
The $*$ operation above is essentially a process in which the convolution kernel (here the shared weight is the convolution kernel) $W_i$ performs convolution operations on all the related feature maps at the ${(l-1})^{th}$ layer, then sums them up and adds a bias parameter $b$ to obtain the final excitation value ${h^{l}_{W,b}(x')}$ at layer $l$ by taking sigmoid.



\subsubsection{The Proposed Recurrent Memory Unit}
\begin{figure}[!htb]
  \centering
  \includegraphics[width=7cm]{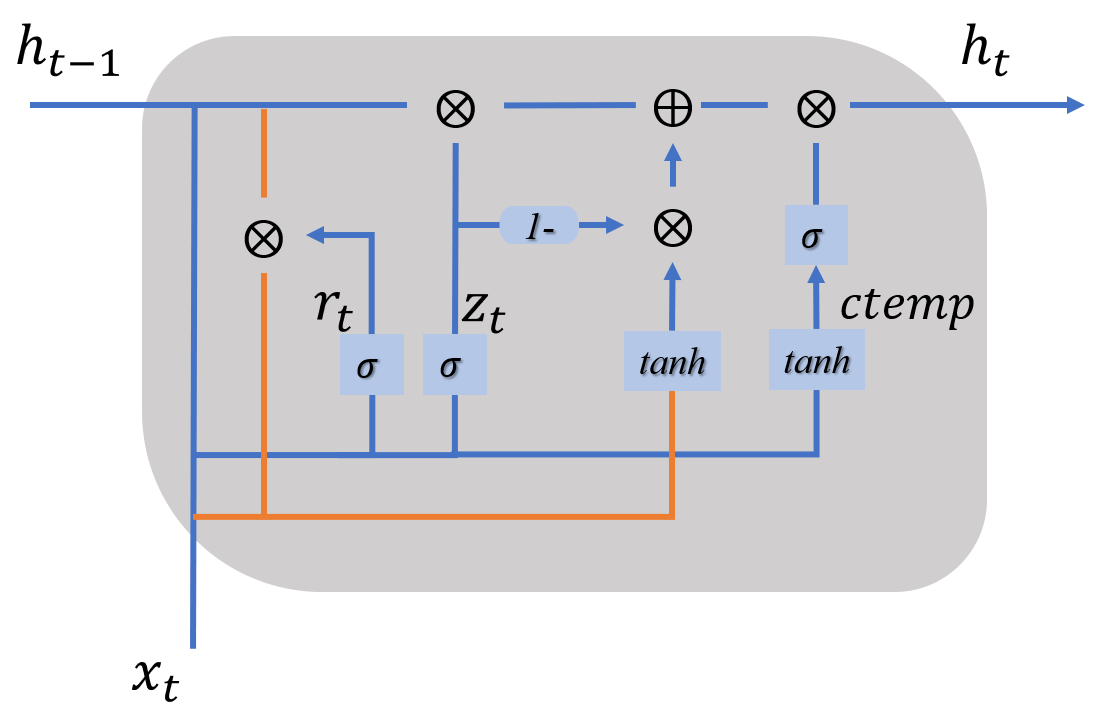}
  \centering
  \caption{The proposed recurrent memory unit. The internal structure and operations are shown in this figure, and $ctemp$ is the new added path in the original GRU cell.}
   \label{grucell}
\end{figure}

We feed the original data into a two-layer memory module to extract temporal information and obtain the corresponding feature matrix. A variant of long short-term memory (LSTM), called the gated recurrent unit (GRU), is proposed to learn long-term dependency information from the input dynamic data, merging the input and forgot gates into an update gate and fusing node and hidden states in each node of the original LSTM, which was first proposed in \cite{Chung2014Empirical}. We improve the performance of the original GRU cell by adding a temporary path and modifying the state settings. Fig. \ref{grucell} shows the internal structure of the proposed recurrent memory unit and illustrates the cooperation of all the gates.


Because of slow convergence of the original GRU in gait data classification, we improve the gating structure in its internal deployment nodes by adding a new path and a temporary state to extract representative features. The improved GRU cell is shown in Fig.\ref{grucell}. The new cell of GRU is illustrated in Eqs. (\ref{z_t})-(\ref{impv_h_t})
\begin{equation}
{z_t} = \sigma({W_z}\cdot[{O_{t-1}},{x'_t}]),
\label{z_t}
\end{equation}
\begin{equation}
{r_t} = \sigma({W_r}\cdot[{O_{t-1}},{x'_t}]),
\label{r_t}
\end{equation}
\begin{equation}
{{\tilde h}_t} = \tanh (W\cdot[{r_t}\odot{O_{t-1}},{x'_t}]),
\label{h'_t}
\end{equation}
where $x'_{t}$ and $O_{t-1}$ are the input and previous hidden states respectively in each memory unit, $\sigma$ and $\tanh$ are the logistic functions. $z_t$ indicates the output of update gate to the network at time step $t\in\{1,2,3...,T\}$, $r_t$ denotes the reset gate. The update gate $z_t$ determines whether or not the hidden state is to be updated with a new hidden state $\tilde h$. The reset gate $r_t$ decides whether or not the previous hidden state is ignored.

\begin{equation}
{ctemp} = \tanh({W_{ctemp}}\cdot[{O_{t-1}},{x'_{t}}]),
\label{impv_ctemp}
\end{equation}
\begin{equation}
{c_t} = (1-{z_t})\odot{{\tilde h}_t}+{z_t} \odot {O_{t-1}},
\label{impv_c_t}
\end{equation}
\begin{equation}
{O_t} = {c_t}\odot \sigma({ctemp}),
\label{impv_h_t}
\end{equation}
where $ctemp$ represents the temporary state to select the information of $x'_{t}$ and $O_{t-1}$ while $c_t$ denotes the final state of the original GRU. It means that we take the advantage of the information selector to extract feature from $c_t$. $O_{t}$ indicates actual activation of the proposed node at time step $t\in\{1,2,3...,T\}$. Experiments show that it can converge faster than the original GRU.

Once having obtained the temporal information from the improved GRU, we use a stacking method to mask the output of each convolution layer with temporal features, forming a three-dimensional cube containing both temporal and spatial features. The representation is shown as follows:

\begin{equation}
{g_{st}(x)} = tile({O_t})+{h^{l}_{W,b}(x')},
\label{g(x)}
\end{equation}
where $g_{st}(x)$ is the function to obtain spatio-temporal features, $tile({O_t})$ is the function to copy the input into multi-dimensional data, which is consistent with the output of ${h^{l}_{W,b}(x')}$, and then add it to ${h^{l}_{W,b}(x')}$ to form a tensor $g$. We use a 3D convolutional kernels/capsules with a step size of 1 to get the output of the primary capsule layer $g'=W_{conv}g$.



\subsection{High-Level Feature Extractor}
After having extracted temporal and spatial features (low-level features), we use capsules to process these features to obtain structural and relationship information (high-level features) of gaits, proposed to solve the problem of information loss on CNN. Different capsules represent different target positions, revealing the relative relationship between different spatial and temporal feature points. Then, according to the relationship between capsules, more important features are selected, and higher-level features are extracted by a dynamic routing algorithm.


In CapsNet, detailed posture information (such as exact locations, rotation, thickness, tilt and size of objects) will be saved in the network before we can restore it. Slight changes in the input will bring small changes to the output, and information will be saved. This allows CapsNet to use a simple and unified architecture for different visual tasks. Therefore, we design a capsule layer as the first layer in the high-level feature extractor.


\subsubsection{Capsule Layer}
The calculation process of the primary capsule layer is a stack of 8 parallel convolutional layers, each of which has a 3D convolutional kernel/capsule with a step size of 1.

The operation of this layer is similar to the convolution that is divided into different capsules. Since the output of the capsule is a vector, a powerful dynamic routing mechanism can be used to ensure that the output is sent to the appropriate parent in the previous layer. Therefore, the structure and direction of the feature are better reflected.

The input vector of the primary capsule layer is equivalent to the scalar input of traditional neural network neurons, and the calculation of the vector is equivalent to the mode of propagation and connection between the primary capsule and digit capsule.

The calculation of input vectors can be divided into two stages, i.e. linear combination and dynamic routing. The linear combination process can be expressed by the following formula:
\begin{equation}
{\hat{g'}_{j|i}} = W_{ij}g'_i,
\label{g'_hat}
\end{equation}
where ${\hat{g'}_{j|i}}$ is a linear combination of $g'_i$, which can be seen as the output of the first layer of neurons in the fully-connected network to a certain neuron in the next layer with different strength connections. ${\hat{g'}_{j|i}}$ represents the output vector of the first capsule of the previous layer and the corresponding weight vector ($W_{ij}$ represents the vector rather than the element). ${\hat{g'}_{j|i}}$ can also be understood as the strength of connecting to $j^{th}$ Capsule of the next layer when the former layer is the Capsule $i$.

After determining ${\hat{g'}_{j|i}}$, we need to use the dynamic routing algorithm to allocate the second stage to calculate the output $s_j$:

\begin{equation}
{s_{j}} = \sum_ic_{ij}{\hat{g'}_{j|i}},
\label{s_j}
\end{equation}
The parameter $c_{ij}$ can be updated iteratively. The input $s_j$ of the next capsule can be obtained by Routing, and then squashing non-linear function can activate the input $s_j$.

By adding the routing mechanism to the capsule, a set of coupling coefficients $c_{ij}$ can be found, which are updated and determined iteratively by dynamic routing process:
\begin{equation}
{c_{ij}} = \frac{exp(b_{ij})}{\sum_k exp(b_{ik})}.
\label{routing}
\end{equation}
which makes the prediction vector ${\hat{g'}_{j|i}}$ consistent with the output vector. $b_{ij}$ depends on the location and type of two capsules, and $c_{ij}$ can be iteratively updated with the consistency of the measurement.

\begin{figure*}[!htb]
 \centering
 \includegraphics[width=11cm]{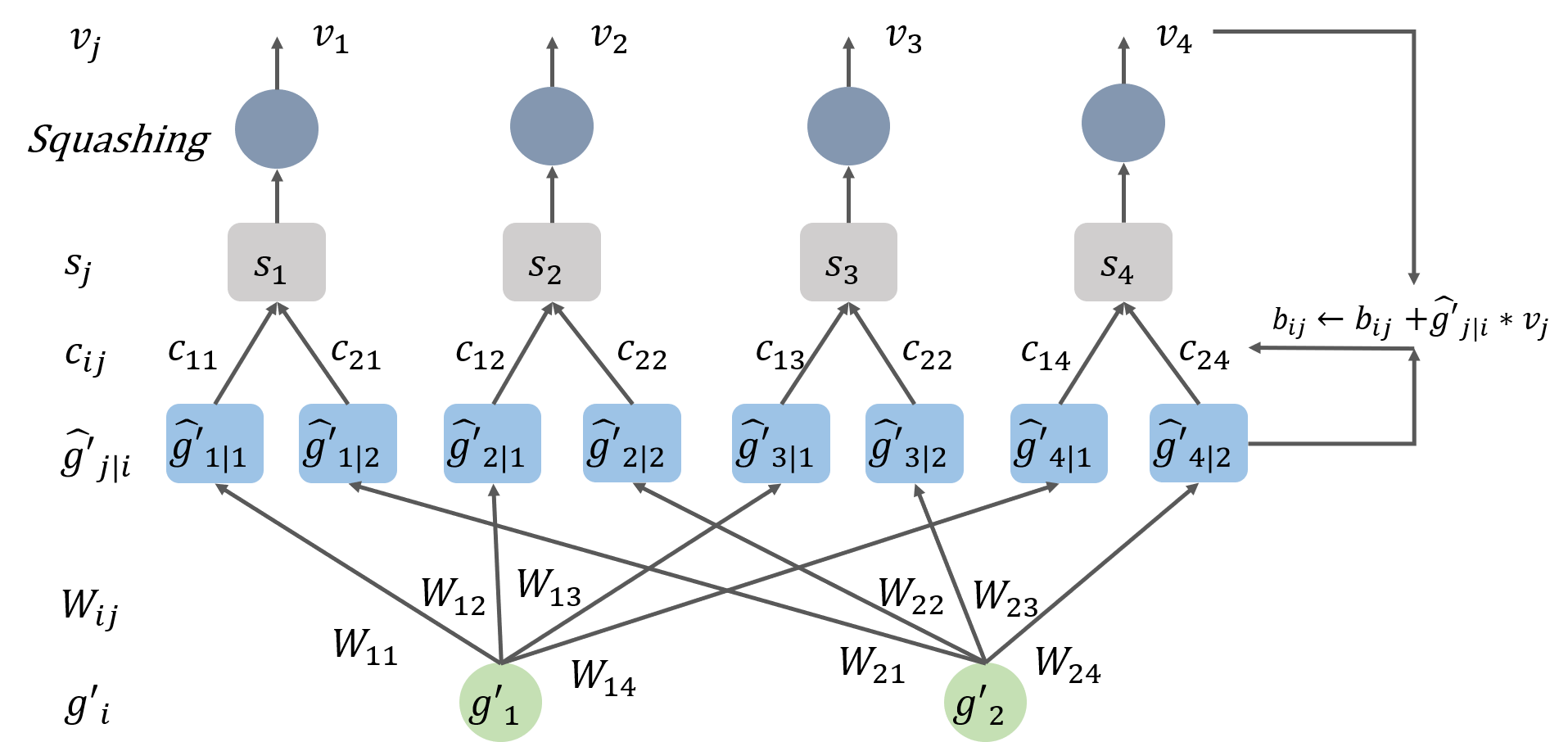}
 \caption{The propagation process between the two capsule layers. There are three steps in this process: 3D convolution ($\hat{g'}_{j|i}$), dynamic routing ($s_j$), and squashing ($v_j$).}
  \label{2capsules}
\end{figure*}



The dissemination and distribution of the whole hierarchy can be divided into two parts. The first part is the linear combination between $g'_i$ and  ${\hat{g'}_{j|i}}$ in the following figure, and the second part is the routing process between ${\hat{g'}_{j|i}}$ and $s_j$. The information transmission and calculation between the two capsule layers are shown in Fig.\ref{2capsules}.

Assuming there are two capsule units ($g'_i$) in the primary capsule layer, and four capsules are passed from this level to the next level $v_j$. $g'_1$ and $g'_2$ are the tensors obtained from the spatio-temporal feature extraction layer, i.e. capsule units containing a group of neurons, which are multiplied by different weights $W_{ij}$ to form $\hat{g'}_{j|i}$, respectively. The predictive vector is then multiplied by the corresponding coupling coefficient $c_{ij}$ and passed to a specific digit capsule unit. The input $s_j$ of different capsule units is the weighted sum of all possible incoming prediction tensors. Then we collect different input tensors $s_j$. By putting the input tensors into the squashing non-linear function, we have the output tensors $v_j$ of the latter capsule unit. We use the product of the output vector $v_j$ and the corresponding prediction vector ${\hat{g'}_{j|i}}$ to update the coupling coefficient $c_{ij}$. Such an iterative update does not need to apply back-propagation.

The activation of neurons in a capsule indicates the properties of specific entities in data. These properties include different parameters, such as gait data position, size, direction, deformation, speed, reflectivity, color, and texture. The length of the input and output tensors represents the probability of the occurrence of a feature, so its value must be between 0 and 1.

In order to achieve this compression and implement the activation function at the capsule level, a non-linear function called ``squashing" ensures that the length of short tensors can be reduced to almost zero, while the length of long tensors can be reduced to near but no more than 1. The following is the expression of the non-linear function:

\begin{equation}
{v_{j}} = \frac{{||s_j||}^2}{1+{||s_j||}^2}\frac{s_j}{||s_j||},
\label{squashing}
\end{equation}
where ${v_j}$ is the output vector of capsule ${j}$, ${s_j}$ is the weighted sum of all the tensors of the capsule output from the previous layer to capsule ${j}$ of the current layer, and $s_j$ is the input vector of capsule $j$. The non-linear function can be divided into two parts. The first part $\frac{{||s_j||}^2}{1+{||s_j||}^2}$ is the scalar of the input vector $s_j$, and the second part $\frac{s_j}{||s_j||}$ is the unit vector of the input vector $s_j$. The non-linear function not only retains the direction of the input vector but also compresses the length of the input vector into the interval [0,1]. The non-linear function can be regarded as compression and reallocation of the length of the vector, so it can also be regarded as a way of activating the output vector after the input vector.

Obtaining the output high-level feature of the digit capsule layer, we use the $softmax$ function for classification. In order to comprehensively consider the effectiveness of feature extraction of the whole network, we classify the output features of each level and construct a Bayesian model to determine the final classification results using those predictions.

\subsubsection{Relationship Layer}
In terms of the primary capsule layer, all the capsules are independent of each other. The extracted features can not only represent the local features (arms, legs, or head) of gait, but also interpret the relationship between local and global features (the whole body). However, the relationship between capsules has been neglected, so we design a capsule relationship layer to preserve the relationship between capsules, that is, the relationship between local features.

In order to preserve the relative position, direction, and state between local features, we use the transfer matrix of the former capsule and the latter capsule to represent their relationship and send them to the $softmax$ classifier. The specific calculation is as follows:
\begin{equation}
{R_{i}} =g'_i*{(g'_{i+1})}^{-1},~~i<32.
\label{Relationship}
\end{equation}

Assuming $R_{i}$ is the relationship matrix of the capsules in the primary capsule layer. $g'_i$ and $g'_{i+1}$ are adjacent capsules. We use the inverse of ${(g'_{i+1})}$ to find the transfer matrix $R_{i}$.

\subsection{Decision-Making Layer}
According to the output of different feature extraction layers, we add four classifiers to evaluate the features of the four layers and output specific class labels. For the classification of the extracted features, the Bayesian model is built for the output of the four classifiers to select more accurate classification results.

\subsubsection{Classification Layers}
Firstly, we embed a softmax multi-classifier on the GRU feature extractor to evaluate whether or not temporal features are beneficial to classification. Furthermore, to make spatial and temporal features more distinct and easy to classify, we add two fully-connected layers to the convolutional layer, followed by a softmax function to classify the result, and integrate the feature cubes to obtain high-level information.

\subsubsection{Bayesian Modeling Process}
After classification, we have four label vectors. We merge these four results into a label matrix, regard the four label values predicted as features, and use the Bayesian model to maximize the posterior probability of class estimation. Assuming that there are $n$ classes, using $c_1$, $c_2$, and $c_n$ to denote, $X={x_1, x_2, ...,x_n}$ represents each sample in a training batch with a certain label $c_i,i\in[1,n]$.
\begin{equation}
{\max P(c_i|X)} = \max \frac{P(X|c_i)P(c_i)}{P(X)}.
\label{bayes}
\end{equation}

Because $P(X)$ is constant for all the classes, the process of maximizing a posteriori probability $P(c_i|X)$ can be transformed into maximizing probability $P (X|c_i)P(c_i)$. It is usually assumed that the attributes are independent of each other, and prior probability $P(x_1|c_i), P (x_2|c_i),..., P(x_n|c_i)$ can be obtained from the training dataset. According to this method, the probability $P(X|c_i)P(c_i)$ that $X$ belongs to each category of $c_i$ can be calculated for a sample $X$ of an unknown category, and then the category with the greatest probability can be selected as its class label.

\subsection{Loss Function}
Combining standard categorical cross-entropy loss of the first three outputs of ASTCapsNet, and the margin loss $l_{dc}$ of the last output of the digit capsule layer, the final joint loss function is formed as:

\begin{equation}\small
\begin{aligned}
l_{dc}=T_c max(0,m^+-||v_c||)^2 + \lambda(1-T_c) max(0,||v_c||-m^-)^2,\\
\end{aligned}
\label{marginloss}
\end{equation}
where $c$ is the class label of a sample, $T_c$ is the indicate function (1 if $c$ exists, else 0), $m^-$ is the top margin of $||v_c||$, $m^+$ is the bottom margin of $||v_c||$.

\begin{equation}\small
\begin{aligned}
{\mathcal{L}(y,y')} & = l_{tp}+l_{st}+l_{pc}+l_{dc}\\
& = -\sum_j{y_{{tp}_j}'}\log(\frac{e^{{y_{tp}}_j}}{\sum_{i=1}^n e^{{y_{tp}}_i}}) -\sum_j{y_{{st}_j}'}\log(\frac{e^{{y_{st}}_j}}{\sum_{i=1}^n e^{{y_{st}}_i}})\\
&-\sum_j{y_{{pc}_j}'}\log(\frac{e^{{y_{pc}}_j}}{\sum_{i=1}^n e^{{y_{pc}}_i}})+\sum_cl_{dc},
\end{aligned}
\label{loss}
\end{equation}
where $l_{tp}$ and $l_{st}$ represent the classification loss of temporal and spatio-temporal data, $l_{pc}$ denotes the classification loss in the primary capsule layer. $l_{dc}$ is the margin loss in the digit capsule layer. $\mathcal{L}(y,y')$ is the joint loss of the four classification results. In the optimization process, we use Adam optimizer to consider the first and second moments of the gradients in order to reduce the loss quickly. $y_{{tp}_j}',y_{{st}_j}',y_{{pc}_j}'$ denote the $j^{th}$ true label of a training batch while $y_{{tp}_j},y_{{st}_j},y_{{pc}_j}$ represents the predicted label.

\section{Experiment}

The experiment was conducted on five different datasets, and the results of the experiment were evaluated against several state-of-the-art methods. The following is a list of all the comparison methods:

\textbf{Classifiers:}
\begin{itemize}
\setlength{\itemsep}{0pt}
\setlength{\parsep}{0pt}
\setlength{\parskip}{0pt}

\item DT (decision tree) refers to a decision support tool using a tree-like model of decisions and their possible consequences.
\item GBDT (gradient boosting decision tree) is an iterative decision tree algorithm, which is composed of multiple decision trees. The results of all the trees are accumulated to determine the final label.
\item LR (logistic regression) is a kind of nonlinear regression models and a machine learning method for probability estimation.
\item RF (random forest) is a classifier with multiple decision trees.
\item KNN (k-nearest neighbor) is a classifier that finds $k$ nearest instance and votes to determine the class name of the new instances.
\end{itemize}

\textbf{Deep Models:}
\begin{itemize}
\setlength{\itemsep}{0pt}
\setlength{\parsep}{0pt}
\setlength{\parskip}{0pt}
\item GRU (gated recurrent unit) is a gating mechanism in recurrent neural networks, which has fewer parameters than LSTM.
\item LSTM (long-short term memory) is a model with expandable nodes that are suitable for temporal data.
\item BiLSTM (bidirectional long-short term memory) is composed of a forward LSTM and a backward LSTM.
\item CapsNet (capsule network) is a deep neural network for dynamic analysis of data structure and spatial information.
\item Attention LSTM is an LSTM model with an attention mechanism.
\item CNN (convolutional neural network) is a network for learning spatial features.
\end{itemize}

Additionally, we also use several gait recognition methods from the literature because they use the dataset mentioned in this paper.

\textbf{Other Studies:}
\begin{itemize}
\setlength{\itemsep}{0pt}
\setlength{\parsep}{0pt}
\setlength{\parskip}{0pt}
\item TL-LSTM\cite{8397466} is a two-layer LSTM for gait recognition.
\item HCF+SVM\cite{Gianaria2019} is a method using hand-crafted features and SVM for classification.
\item GCP\cite{Hong2017A} is a gait cycle partitioning method.
\item WT+GA\cite{Shi2015A} is the combination of wavelet transform and a genetic algorithm.
\item DT+RF\cite{8713735} is the combination of distance transform and random forests.
\item SD\cite{7532940} (static and dynamic feature extraction method) is a human walking model including the static and dynamic gait features.
\item FLM\cite{8653351} (frame-level matching) is a frame-level matching method for gait recognition.
\item RBF\cite{ZENG2015246} (radial basis function) is a neural network for gait analysis and recognition.
\item DCLSTM\cite{Zhao2018Dual} (dual-channel LSTM) is a temporal LSTM model for multimodal gait recognition.
\item Q-BTDNN\cite{NANCYJANE2016169} (Q-backpropagated time-delay neural network) is presented to identify the gait disturbances in Parkinson's disease.
\item 2D-CNN+LSTM\cite{8781511} is the combination of 2D-CNN and LSTM for gait spatio-temporal information extraction.
\item SFM \cite{10.5555/3305381.3305543} (state-frequency memory) is a recurrent network that allows separating dynamic patterns across different frequency components and their impacts on modeling the temporal contexts of the sequences.
\item CapProNet\cite{NIPS2018_7823} (capsule projection network) can learn an orthogonal projection matrix for each capsule subspace to classify different objects.
\end{itemize}

\begin{figure*}
     \centering
    \begin{subfigure}[b]{0.18\textwidth}
         \centering
         \includegraphics[width=3.6cm,height=3.3cm]{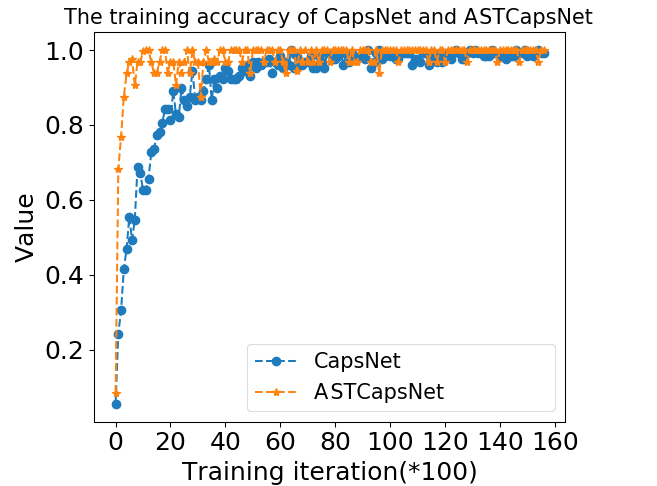}
         \caption{UNITO}
         \label{train_acc:unito}
     \end{subfigure}
     \begin{subfigure}[b]{0.18\textwidth}
         \centering
         \includegraphics[width=3.6cm,height=3.5cm]{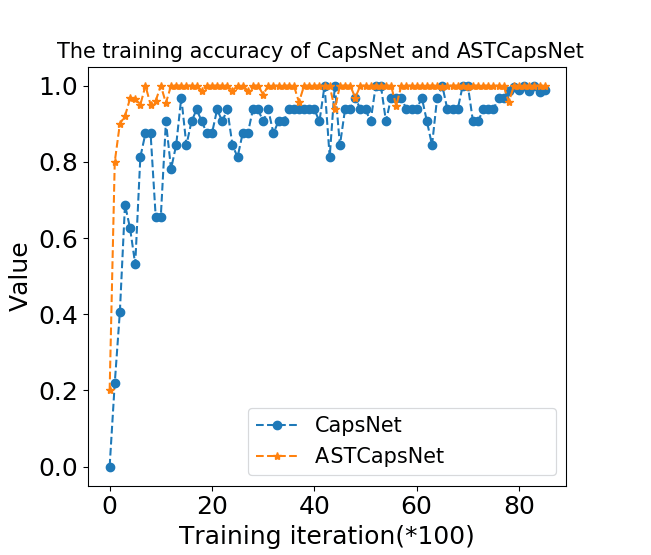}
         \caption{CASIA-A}
         \label{train_acc:casia}
     \end{subfigure}
     \begin{subfigure}[b]{0.18\textwidth}
         \centering
         \includegraphics[width=3.6cm,height=3.3cm]{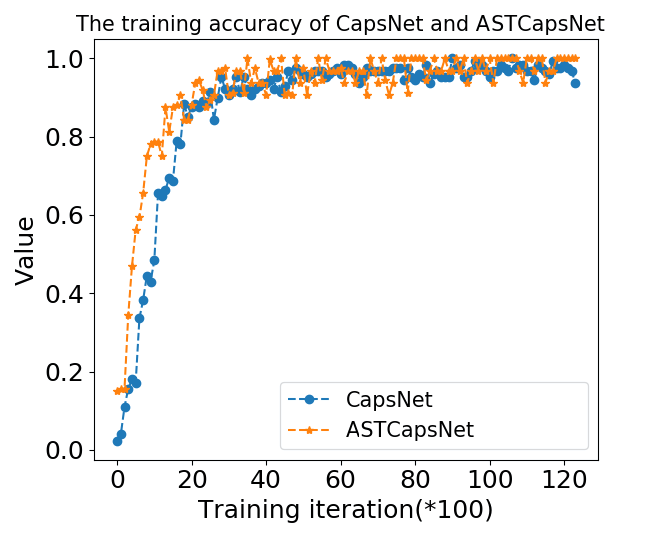}
         \caption{SDUgait}
         \label{train_acc:sdu}
     \end{subfigure}
     \begin{subfigure}[b]{0.18\textwidth}
         \centering
         \includegraphics[width=3.6cm,height=3.5cm]{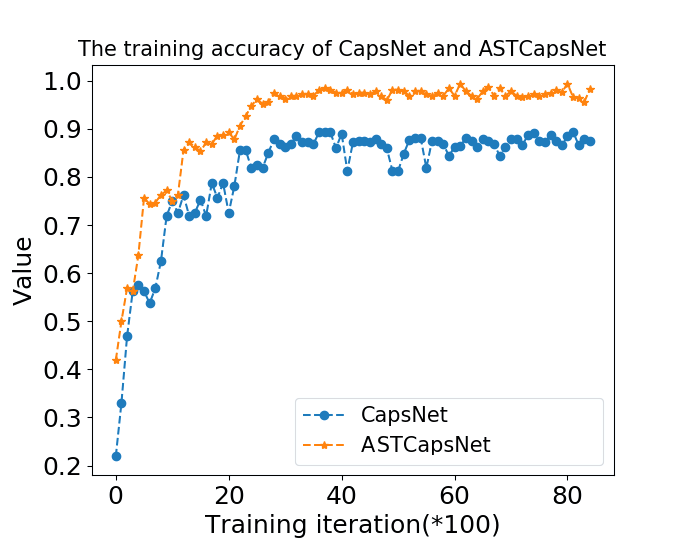}
         \caption{NDDs}
         \label{train_acc:ndds}
     \end{subfigure}
     \begin{subfigure}[b]{0.18\textwidth}
         \centering
         \includegraphics[width=3.6cm,height=3.5cm]{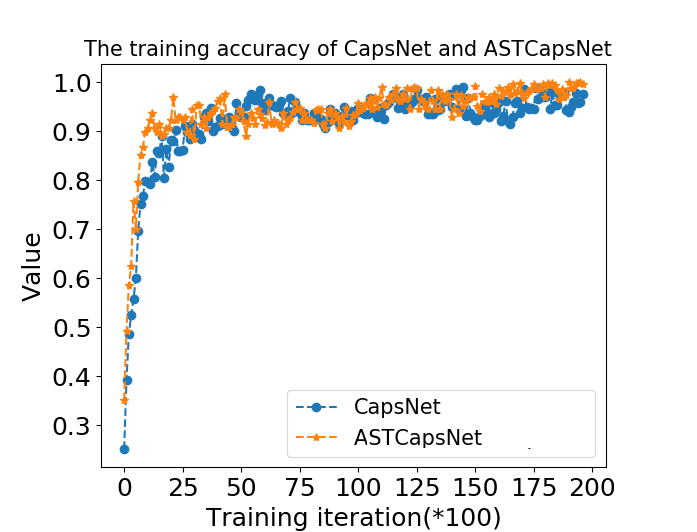}
         \caption{PD}
         \label{train_acc:ga}
     \end{subfigure}
\caption{The training accuracy of the original CapsNet and ASTCapsNet on the five datasets.}
\label{train_acc}
\end{figure*}

\begin{figure*}
     \centering
    \begin{subfigure}[b]{0.18\textwidth}
         \centering
         \includegraphics[width=3.5cm,height=3.5cm]{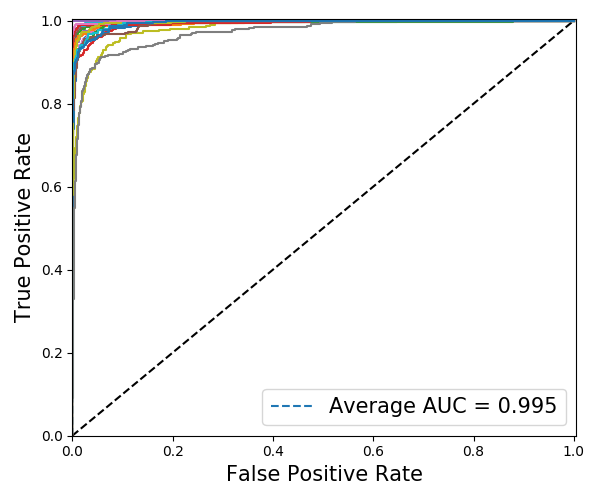}
         \caption{UNITO}
         \label{roc:unito}
     \end{subfigure}
     \begin{subfigure}[b]{0.18\textwidth}
         \centering
         \includegraphics[width=3.5cm,height=3.5cm]{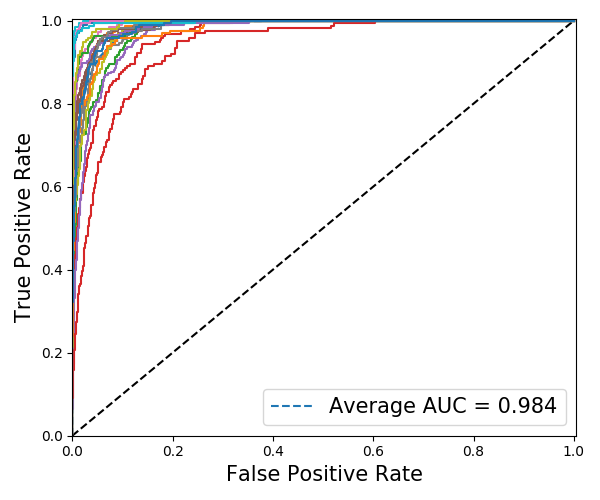}
         \caption{CASIA-A}
         \label{roc:casia}
     \end{subfigure}
     \begin{subfigure}[b]{0.18\textwidth}
         \centering
         \includegraphics[width=3.5cm,height=3.5cm]{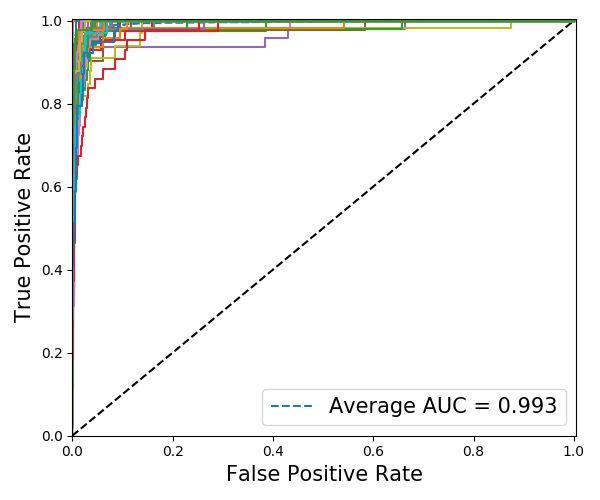}
         \caption{SDUgait}
         \label{roc:sdu}
     \end{subfigure}
     \begin{subfigure}[b]{0.18\textwidth}
         \centering
         \includegraphics[width=3.5cm,height=3.5cm]{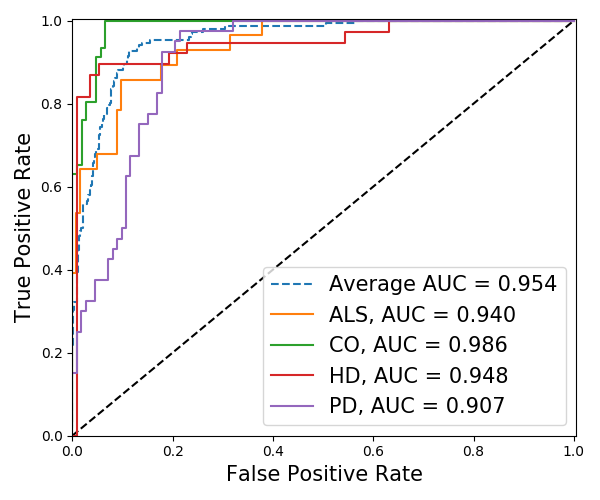}
         \caption{NDDs}
         \label{roc:ndds}
     \end{subfigure}
     \begin{subfigure}[b]{0.18\textwidth}
         \centering
         \includegraphics[width=3.5cm,height=3.5cm]{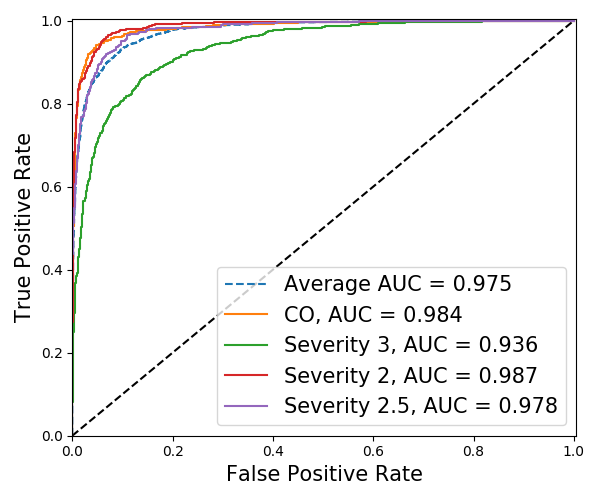}
         \caption{PD}
         \label{roc:ga}
     \end{subfigure}
\caption{ROC curves of ASTCapsNet on the five datasets. The legend shows the AUC values of different classes in the dataset, and there are too many classes of UNITO, CASIA-A, and SDUgait datasets to be fully represented by the legend, so we only give average AUC values. Each line represents the AUC value of the corresponding class. ASTCapsNet has a high recognition rate of each class and the curve is concentrated in the upper left corner. Besides, the micro-average AUC values are compared, which shows the strong learning ability of ASTCapsNet.}
\label{roc}
\end{figure*}

\subsection{Experimental Setup}
First, we introduce the datasets used in our experiments and then describe the training details.

\subsubsection{Datasets}
We perform experiments on three gait datasets (normal people), including CASIA Gait dataset \cite{CASIA}, UNITO dataset \cite{gianaria2014human}, and SDUgait dataset \cite{SDUgait}, and also verify their extensibility on two gait datasets (neurodegenerative patients) - NDDs dataset \cite{physionet-ndd} and PD dataset \cite{physionet-pd}. These datasets are collected using different sensors, such as cameras, infrared sensors, and force sensors. The data types include human skeletons, gait contour maps, and so on. We will introduce these datasets in detail.

\textbf{UNITO Dataset:} This includes skeleton data of 20 subjects, acquired with Kinect For Windows SDK 1.x. The subjects were asked to walk naturally along a corridor, towards the camera (FRONT view) and away from the camera (REAR view), for 10 times, with a total of 20 gait sequences for each subject (10 FRONT and 10 REAR). The order of the 20 joints is the same as the Kinect Skeleton Map. For each joint, 3D coordinates of subjects are recorded.

\textbf{CASIA Gait Dataset:} In this dataset, we choose Dataset A (former NLPR Gait Dataset), which includes 20 persons. Each person has 12 preprocessed gait contour sequences, 4 sequences for each of the three directions, i.e. parallel, 45 degrees, and 90 degrees to the image plane. The length of the sequences is not identical for the variation of the walker's speed, but it must range from 37 to 127.

\textbf{SDUgait dataset:} The dataset includes 52 subjects, 28 males and 24 females with an average age of 22. Each subject has 20 sequences with at least 6 fixed walking directions and 2 arbitrary directions, and a total of 1040 sequences; The dataset is based on the second-generation Kinect (Kinect V2). The Kinect V2 does not only have a broader view angle but also can produce higher-resolution of depth images with 25 body joints.

\textbf{NDDs dataset:} The dataset contains the gait signals of 48 patients with NDDs (ALS, HD, and PD) and 16 healthy controls (CO). Participants were walking at their usual pace along a 77-m-long hallway for 5 minutes including the recorded signal of stride, swing, and stand times for each leg and double support signals for both legs. An expert physician labeled patients' states from 0 to 13 (0 equal to the most severe state and 13 for a healthy one). To measure time intervals, a 12-bit onboard analog-to-digital converter samples the output of foot switches at 300 Hz \cite{Hausdorff1998Gait}.


\textbf{PD Dataset:} In this study, we have utilized gait signals from PhysioNet. The dataset consists of three PD gait sub-datasets, which are contributed by three researchers (Ga \cite{yogev2005dual}, Ju \cite{hausdorff2007rhythmic} and Si \cite{Frenkel2005Treadmill}). We only select the subset of Ga \cite{yogev2005dual} as an example of PD severity classification.

The dataset includes gait information from 93 patients with the idiopathic PD and 73 healthy controls (average age 66.3, 55\% male). Every participant was asked to walk at their usual, self-selected pace for about two minutes while wearing a pair of shoes with 8 force sensors located under each foot. The sensors measure vertical ground reaction force (VGRF, in Newton) as a function of time at 100 samples per second. The dataset also contains the specific situation of each participant including gender, age, height, weight, and severity level of PD. The PD severity level is graded according to two scales (H\&Y, UPDRS).

\subsubsection{Training Details}
We employ different training settings on these datasets. For the UNITO Dataset, we use 5*5 and 3*3 convolutional kernels in the convolutional layer and primary capsule layer, 7-time steps in LSTM with the batch size of 128, and input dimension of 9 (63D, 21 joints, 3D coordinates), the dropout in LSTM is set to 0.5 to avoid over-fitting.

For CASIA Gait Dataset, we apply all of three directions of the gait contour images and convert the image into text sequences. After resizing the original image to 50 by 50 pixels, the gait data is fed into ASTCapsNet with 9*9 convolutional kernels in the convolutional layer and the primary capsule layer. The time step and input vector are set to 50 and 50 to match the image size. The learning rate is 0.001 in LSTM and other components. The input and time steps in LSTM are 50 and 50 respectively.

For the SDUgait dataset, we only use the skeleton data of the gait, which includes the 3D positions of 21 joints. Differently, the time step in LSTM is set to 7, and the input is 9D (63D, 3*21), that is, the dimension of the input in LSTM is 63. 5*5 and 3*3 convolutional kernels can be used in the convolutional layer and the primary capsule layer respectively.

For the NDDs dataset, the input matrix of a training sample is 12*10 that the input is 12D and the time step is 10. The size of the convolutional kernels is set to 5*5 based on the scale of the input data. The hidden output of LSTM is 128D to represent the temporal feature of the gait.

For the PD Dataset, the size of a training sample is 19*100, which indicates that the input is 19D and the time step is 100. The size of the convolutional kernels is set to 9*9 based on the scale of the input data. The hidden output of LSTM is 128D to represent the temporal feature of gait.

The parameters of other layers are set according to different datasets and fine-tuned to achieve good performance. The model is implemented with Tensorflow.

\subsection{Experiments on Normal Gait Datasets}
In this subsection, we evaluate the ASTCapsNet model on normal gait datasets for identity recognition: UNITO, CASIA and SDUgait datasets.

\subsubsection{Experiments on the UNITO Skeleton Dataset}
 \begin{figure}[htp]
\setlength{\abovecaptionskip}{0.cm}
\setlength{\belowcaptionskip}{-0.cm}
  \centering
  \includegraphics[width=9cm]{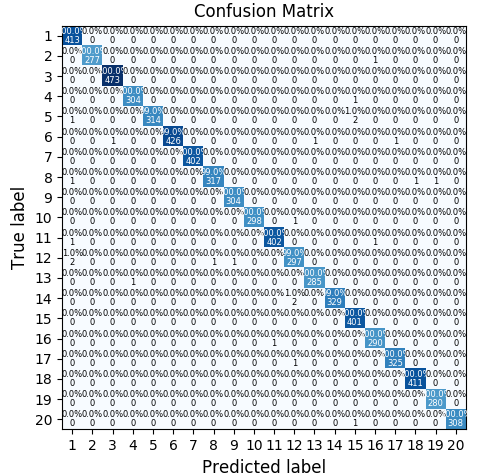}
  \caption{The confusion matrix of classification results on the UNITO Skeleton Dataset.}
  \label{unito_confusion_matrix}
\end{figure}

\begin{table}[!htb]
\centering 
\caption{Results on the UNITO skeleton dataset.}
\scalebox{0.8}{
\begin{tabular}{lc|lc|lc}
  \hline
  \hline
  Classifiers&Acc&Deep Models&Acc&Others&Acc\\
  \hline
  DT&87.29\%& GRU&97.95\%&TL-LSTM\cite{8397466}&97.33\%\\
  GBDT&90.82\%&LSTM&95.44\%&HCF+SVM\cite{Gianaria2019}&97.00\%\\
  LR&67.51\%&BiLSTM&96.45\%&SFM\cite{10.5555/3305381.3305543}&96.97\%\\
  RF&96.47\%&CapsNet&97.33\%&CapProNet\cite{NIPS2018_7823}&98.69\%\\
  KNN&90.00\%&Attention LSTM&97.92\%&&\\
  &&CNN&95.84\%&&\\
 \hline
  ASTCapsNet&\textbf{\textit{99.65\%}}\\
\hline
\hline
\end{tabular}}
  \label{tb_UNITO}
\end{table}

This experiment aims to verify that our proposed model can perform well in the three-dimensional skeletal dataset, which outperforms the other methods on this dataset.

Firstly, in this dataset, we have 20 subjects for classification, and use 80\% of the data for training and 20\% for testing. The performance of the proposed model is shown in Fig. \ref{unito_confusion_matrix}. It is obvious that the gait recognition rate of the model is higher than 99.0\% for each of 20 subjects, and we finally achieve the test result of 99.65\%. 

Furthermore, we compare our model with several advanced algorithms for this dataset in recent years, which are demonstrated in Table \ref{tb_UNITO}. In the light of the results of five traditional classifiers, i.e., decision tree (DT), gradient boosting decision tree (GBDT), logic regression (LR), random forest (RF), k-nearest neighbor (KNN), and temporal models, i.e., gated recurrent unit (GRU), long-short term memory (LSTM), attention LSTM and bi-directional long short-term memory (BiLSTM), we can see that the average result of the temporal model is higher than that of the traditional classifiers, and the performance of RF and attention LSTM is the best among the models.

For other studies, Gianaria \textit{et al.}, \cite{Gianaria2019} uses handcrafted features as input to the SVM classifier and achieves the result of 97\%, which needs to calculate the relative distance between joints (FoRD) and the sway of joints (FoS). Li \textit{et al.}, \cite{8397466} apply a two-layer LSTM in this dataset and obtain the result of 97.33\% which considers the temporal changes of gait. The performance of SFM \cite{10.5555/3305381.3305543} exceeds LSTM and BiLSTM, indicating the advantages of using state-frequency components. Although the original CapProNet\cite{NIPS2018_7823} can perform better than these methods by applying capsule subspace, our model achieves the highest accuracy among them.

Additionally, we also show the training accuracy of the original CapsNet and ASTCapsNet in Fig.\ref{train_acc:unito}. We can see that ASTCapsNet converges faster than the original CapsNet, and the final training accuracy is closer to 100\%. Fig.\ref{roc:unito} shows the receiver operating characteristic curve (ROC) for the evaluation of the proposed model. The curve is close to the upper left corner, and the average area under curve (AUC) value approaches to 1, indicating ASTCapsNet achieves satisfactory performance.

\subsubsection{Experiments on the CASIA Gait Dataset}

\begin{table}[!htb]
\centering 
\caption{Results on the CASIA gait dataset.}
\scalebox{0.8}{
\begin{tabular}{lc|lc|lc}
  \hline
  \hline
  Classifiers&Acc&Deep Models&Acc&Others&Acc\\
  \hline
  DT&67.78\%&GRU&96.97\%&GCP\cite{Hong2017A}&92.50\%\\
  GBDT&70.65\%&LSTM&96.73\%&WT+GA\cite{Shi2015A}&97.32\%\\
  LR&86.49\%&BiLSTM&97.03\%&DT+RF\cite{8713735}&99.60\%\\
  RF&73.32\%&CapsNet&93.91\%&SFM\cite{10.5555/3305381.3305543}&97.51\%\\
  KNN&70.71\%&Attention LSTM&96.79\%&CapProNet\cite{NIPS2018_7823}&95.49\%\\
  &&CNN&90.83\%&&\\
 \hline
  ASTCapsNet&\textbf{\textit{99.69\%}}\\
\hline
\hline
\end{tabular}}
  \label{tb_casia}
\end{table}

This experiment aims to verify that our proposed model can perform well in the 2D silhouette image dataset, which also outperforms the other methods for this dataset.

\begin{figure}[htp]
\setlength{\abovecaptionskip}{0.cm}
\setlength{\belowcaptionskip}{-0.cm}
  \centering
  \includegraphics[width=9cm]{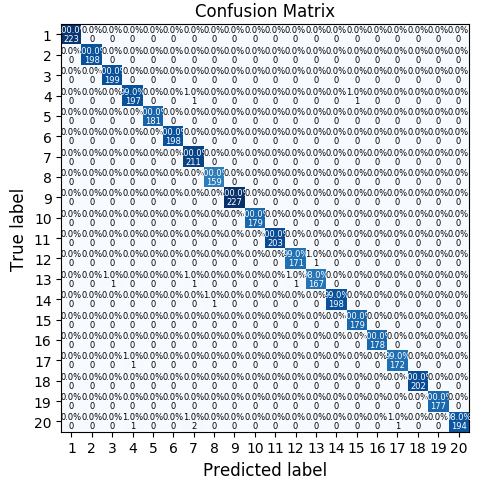}
  \caption{The confusion matrix of classification results on the CASIA-A Gait Dataset.}
  \label{casia_confusion_matrix}
\end{figure}

In this dataset that includes 20 subjects for identification, we use 80\% of the data for training and 20\% for testing. The performance of the proposed model is shown in Fig. \ref{casia_confusion_matrix}. The correct number and accuracy of the samples on the diagonal line show that the model can accurately determine the identity via gait.

Beyond that, we also compare the experimental results in the past five years, and the model presented in this paper also stands out, which are illustrated in Table \ref{tb_casia}. Because of the temporality of the gait data, the temporal model is more suitable for learning, which achieves better results. Hong \textit{et al.}, \cite{Hong2017A} use a gait cycle partitioning method for gait representation. Distance transform and random forest are also compared for this task \cite{8713735}. \cite{Shi2015A} propose to use wavelet transform and a genetic algorithm for identification, and the result is also shown in Table \ref{tb_casia}. In general, the temporal models are better than the other models, where capsule networks, i.e., CapsNet and CapProNet can effectively analyze the structural and spatial information of gait.

Fig.\ref{train_acc:casia} shows the training accuracy of the original CapsNet and ASTCapsNet on the CASIA-A Dataset. After the $5^{th}$ training iteration, ASTCapsNet achieves higher accuracy than the original CapsNet and quickly converges to nearly 100\%. The average AUC value of the CASIA-A dataset is 98.4\%, the true positive rate (TPR) approaches 1 when the false positive rate (FPR) is 0.3, which verifies the effectiveness of the ASTCapsNet (Fig. \ref{roc:casia}).

\subsubsection{Experiments on the SDUGait Dataset}
\begin{figure}[htp]
\setlength{\abovecaptionskip}{0.cm}
\setlength{\belowcaptionskip}{-0.cm}
  \centering
  \includegraphics[width=7cm]{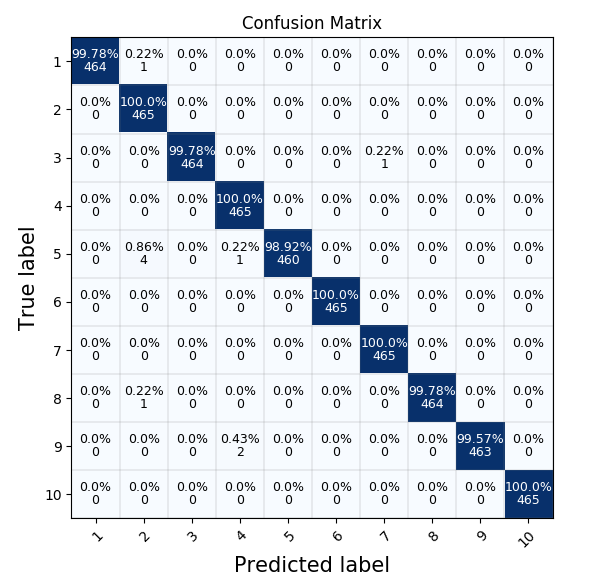}
  \caption{The confusion matrix of the classification results on the SDUGait Dataset. Since the recognition results of 52 subjects cannot be completely shown due to the page size, we select the subjects labeled 1-10 to show the recognition results which is 99.13\%.}
  \label{sdu_confusion_matrix}
\end{figure}

\begin{table}[!htb]
\centering 
\caption{Results on the SDUGait gait dataset.}
\scalebox{0.8}{
\begin{tabular}{lc|lc|lc}
  \hline
  \hline
  Classifiers&Acc&Deep Models&Acc&Others&Acc\\
  \hline
  DT&80.33\%&GRU&95.30\%&SD\cite{7532940}&94.23\%\\
  GBDT&85.02\%&LSTM&94.80\%&FLM\cite{8653351}&98.08\%\\
  LR&63.08\%&BiLSTM&96.38\%&SFM\cite{10.5555/3305381.3305543}&95.02\%\\
  RF&94.26\%&CapsNet&97.00\%&CapProNet\cite{NIPS2018_7823}&97.64\%\\
  KNN&85.28\%&Attention LSTM&96.33\%&&\\
  &&CNN&92.85\%&&\\
 \hline
  ASTCapsNet&\textbf{\textit{99.13\%}}\\
\hline
\hline
\end{tabular}}
  \label{tb_sdu}
\end{table}

This experiment aims to verify that our proposed model can perform well in the 3D skeletal dataset, which also outperforms the other methods applied to this dataset.

Firstly, in this dataset that contains 52 subjects for classification, we use 80\% of the data for training and 20\% for testing. The performance of the proposed model is shown in Fig. \ref{sdu_confusion_matrix}. There are so many categories that the recognition results of each category are not shown and the outline of the color confusion matrix is displayed. We can also see that the recognition error rate of the model for different gaits is very small.

Moreover, we compare this result with some advanced algorithms for this dataset, which are shown in Table \ref{tb_sdu}. The average result of the temporal model is much higher than that of the traditional classifier, and RF and BiLSTM perform the best. Wang \textit{et al.}, \cite{7532940} obtains the classification result of 94.23\% on this dataset by using the static and dynamic feature extraction methods, being view-invariant for gait recognition. Although we only use the 3D coordinates of the skeletal joints in the SDUGait dataset, the proposed model can also well learn the gait patterns. Besides, Choi \textit{et al.}, \cite{8653351} propose a robust frame-level matching method for gait recognition that minimizes the influence of noisy patterns as well as secures the frame-level discriminative power and achieves accuracy of 98.08\%, little lower than ASTCapsNet. The SFM \cite{10.5555/3305381.3305543} here does not show outstanding advantages due to the sparsity of gait data and unobvious frequency variation. The result of CapProNet\cite{NIPS2018_7823} is still slightly higher than that of CapsNet.


Furthermore, we also show the training accuracy of the original CapsNet and ASTCapsNet in Fig. \ref{train_acc:sdu}. The curves of the training accuracy on this dataset are similar. In the $20^{th}$ training iteration, ASTCapsNet begins to be slightly higher than CapsNet until the end of the training, indicating the stability of the model. Fig. \ref{roc:sdu} shows the ROC curve of the proposed ASTCapsNet. We can see that except for one subject with a high detection error rate, the AUC of the other subjects is more than 97\%, reflecting the high performance of the model.

\subsection{Experiments on Abnormal Gait Datasets}
In this subsection, we evaluate the ASTCapsNet model on the abnormal gait datasets for disease classification and diagnosis: NDDs and PD datasets.

\subsubsection{Experiments on the NDDs Dataset}
\begin{figure}[htp]
  \centering
  \includegraphics[width=6cm]{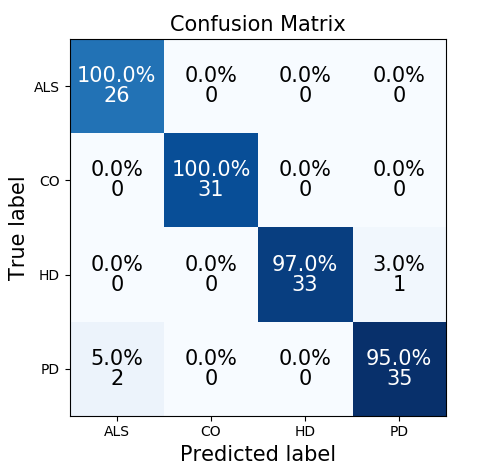}
  \caption{The confusion matrix of classification results on the NDDs Dataset.}
  \label{ndds_confusion_matrix}
\end{figure}

\begin{table}[!htb]
\centering 
\caption{Results on the NDDs dataset.}
\scalebox{0.8}{
\begin{tabular}{lc|lc|lc}
  \hline
  \hline
  Classifiers&Acc&Deep Models&Acc&Others&Acc\\
  \hline
  DT&49.34\%&GRU&92.76\%&RBF\cite{ZENG2015246}&93.75\%\\
  GBDT&67.11\%&LSTM&92.11\%&DCLSTM\cite{Zhao2018Dual}&95.67\%\\
  LR&51.32\%&BiLSTM&93.82\%&SFM\cite{10.5555/3305381.3305543}&93.59\%\\
  RF&65.13\%&CapsNet&88.86\%&CapProNet\cite{NIPS2018_7823}&92.11\%\\
  KNN&73.03\%&Attention LSTM&93.41\%&&\\
  &&CNN&85.58\%&&\\
 \hline
  ASTCapsNet&\textbf{\textit{95.78\%}}\\
\hline
\hline
\end{tabular}}
  \label{tb_ndds}
\end{table}

In this Dataset, our model is used to identify 3 diseases and 1 healthy control group (ALS, HD, PD, and CO). The performance is explained in Fig. \ref{ndds_confusion_matrix}. CO and ALS obtain the result of 100\% while two samples of PD are misclassified into ALS.

Moreover, there are several comparative experiments implemented on this dataset, as illustrated in Table \ref{tb_ndds}. The five classifiers can not provide a good reference because of the small amount of data, and the time series model can capture the dynamic changes of the gait data by more than 90\%. It can be seen that the effect of CapsNet on this dataset is under 90\%, showing a slightly weaker classification ability than CapProNet \cite{NIPS2018_7823}. The effect of the two-channel LSTM using two different sensor data is only inferior to that of ASTCapsNet.

Apart from this observation, we also show the training accuracy of the original CapsNet and ASTCapsNet in Fig. \ref{train_acc:ndds}. The proposed model is approximately 10\% higher than the original CapsNet because the temporal LSTM and Bayesian model are added to compensate for the loss of the features caused by CapsNet. Moreover, the small amount of data does not allow the original CapsNet to learn well. As shown in Fig. \ref{roc:ndds}, the ROC curves of CO tend to be 1, while the curves of ALS, PD, and HD slightly deviate, indicating that the model is more sensitive to the classification of CO due to the big difference between healthy people and patients.


\subsubsection{Experiments on the PD Dataset}
\begin{figure}[htp]
  \centering
  \includegraphics[width=6.8cm]{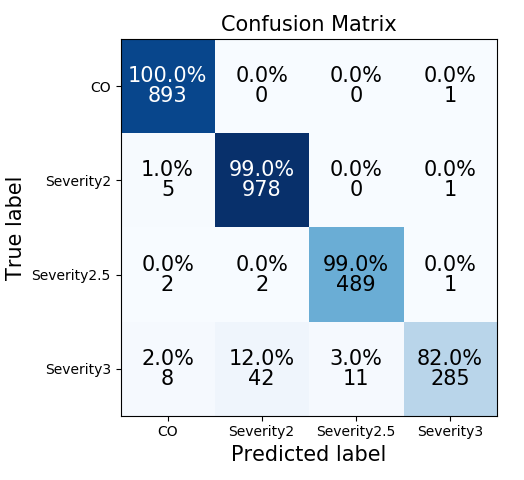}
  \caption{The confusion matrix of classification results on the PD Dataset.}
  \label{ga_confusion_matrix}
\end{figure}

\begin{table}[!htb]
\centering 
\caption{Results on the PD dataset.}
\scalebox{0.8}{
\begin{tabular}{lc|lc|lc}
  \hline
  \hline
  Classifiers&Acc&Deep Models&Acc&Others&Acc\\
  \hline
  DT&77.90\%&GRU&94.89\%&Q-BTDNN\cite{NANCYJANE2016169}&93.10\%\\
  GBDT&84.48\%&LSTM&93.95\%&2D-CNN+LSTM\cite{8781511}&96.00\%\\
  LR&57.23\%&BiLSTM&94.78\%&SFM\cite{10.5555/3305381.3305543}&93.63\%\\
  RF&91.84\%&CapsNet&95.30\%&CapProNet\cite{NIPS2018_7823}&96.58\%\\
  KNN&92.02\%&Attention LSTM&93.53\%&&\\
  &&CNN&89.76\%&&\\
 \hline
  ASTCapsNet&\textbf{\textit{97.31\%}}\\
\hline
\hline
\end{tabular}}
  \label{tb_pd}
\end{table}

In the PD Dataset, we only use the Ga \cite{yogev2005dual} subset as an example to classify the severity of Parkinson's disease based on the gait performance. There are three categories of severity according to the criteria of  H \& Y scale (2=Bilateral involvement without impairment of balance, 2.5=Mild bilateral disease with recovery on the pull test, 3=Mild to moderate bilateral disease; some postural instability; physically independent), and the four categories are compared with the healthy control group.

After training ASTCapsNet, the performance is shown in Fig. \ref{ga_confusion_matrix}. The model is 100\% for healthy people and 82\% for patients with severity 3. We witness that the model does not distinctly extract patient features for severity 3, which leads to misclassification.

Compared with other experiments on this dataset, our proposed model achieves higher classification accuracy. The details are illustrated in Table \ref{tb_pd}. The results of RF and KNN are better, and GRU also shows advantages in temporal models. Q-BTDNN (Q-backpropagated time-delay neural network) is designed for severity classification which can diagnose the gait disturbances in Parkinson's disease. Another study \cite{8781511} uses both 2D-CNN and LSTM to obtain a promising result of 96.00\%, while CapProNet \cite{NIPS2018_7823} performs better. ASTCapsNet still outperforms the original CapsNet and CapProNet \cite{NIPS2018_7823}.

Additionally, we also show the training accuracy of the original CapsNet and ASTCapsNet in Fig. \ref{train_acc:ga}. We can see that after the $125^{th}$ training iteration, ASTCapsNet begins to achieve slightly higher training accuracy than CapsNet until the end of the iteration. The AUC value of the four-class classification for Parkinson's disease is 97.5\%, a high level of recognition (\ref{roc:ga}). The AUC value of severity 3 is low, which means that the boundary between severity 3 and other severity degrees is not clear, and the difference is minor.


\subsection{Performance of the Whole Model}
In this section, we evaluate and compare the performance of the whole model and its internal components to verify the model's ability to extract and classify features.

\subsubsection{Performance of Split Components}
First, we provide the recognition results of each module in Table \ref{tb_components}.  The performance of six individual components is demonstrated on the five datasets respectively.  As shown in Fig. \ref{capsnet_process}, the low-level feature extractor includes the memory and convolution modules for spatio-temporal feature extraction. The high-level feature extractor includes the capsule and capsule relationship layers for relationship and structure feature extraction. ``No relationship layer" is the output of ASTCapsNet after we have removed the relationship layer, and ``no memory module" is the output of ASTCapsNet after removing the memory module. ``Bayesian" model is the original Bayesian model, and ``Bayesian model (last layer)" is the total output of ASTCapsNet.

It can be seen that the average output of the high-level feature extractor is better than that of the low-level feature extractor on these five datasets, which reflects the superiority of the module integration scheme. The performance of the ``no relationship layer" and the ``no memory module" is similar, depending on the fitting degree of the model and data. However, the data does not conform to the Gaussian distribution, leading to the poor outcome of using the Bayesian model directly, and the output of the softmax layer conforms to the Gaussian distribution so that the Bayesian model can be placed in the last layer to retain higher accuracy. Briefly, in the comparison of all the individual modules, the high-level feature extractor makes the most, followed by the memory module, and then the relationship layer.

\begin{table*}[!htb]
\centering
\caption{Performance of individual components in ASTCapsNet.}
\scalebox{1}{
\begin{tabular}{lccccccc}
  \hline
  \hline
 Dataset&Low-Level&High-Level&No Relationship&No Memory&Bayesian Model&Bayesian Model\\
 &Feature Extractor&Feature Extractor&Layer&Module&(Raw Data)&(Last Layer)\\
  \hline
  \hline
  UNITO&95.87\%&99.56\%&97.41\%&97.10\%&77.54\%&\textbf{\textit{99.65\%}}\\
  CASIA-A&97.32\%&99.23\%&97.62\%&95.73\%&73.29\%&\textbf{\textit{99.69\%}}\\
  SDUgait&95.33\%&98.56\%&96.50\%&97.94\%&78.56\%&\textbf{\textit{99.13\%}}\\
  NDDs&93.64\%&95.33\%&94.14\%&90.39\%&60.85\%&\textbf{\textit{95.78\%}}\\
  PD&94.37\%&96.38\%&95.72\%&95.87\%&69.78\%&\textbf{\textit{97.31\%}}\\
\hline
\hline
\end{tabular}}
  \label{tb_components}
\end{table*}

\subsubsection{Separability of the Extracted Features}
\begin{figure*}
     \centering
     \begin{subfigure}[b]{0.19\textwidth}
         \centering
         \includegraphics[width=3.5cm,height=3.5cm]{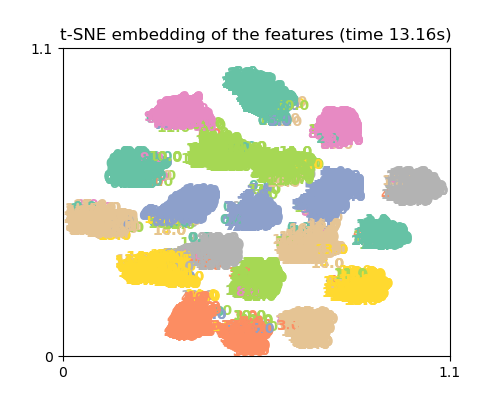}
         \caption{Low-Level Features \\(UNITO)}
         \label{tsne:UNITO1}
     \end{subfigure}
     \begin{subfigure}[b]{0.19\textwidth}
         \centering
         \includegraphics[width=3.5cm,height=3.5cm]{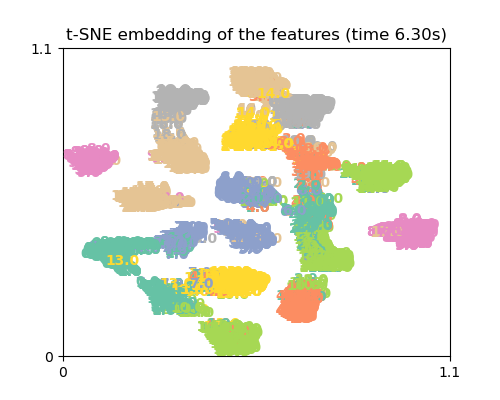}
         \caption{Low-Level Features \\(CASIA-A)}
         \label{tsne:CASIA-A1}
     \end{subfigure}
    \begin{subfigure}[b]{0.19\textwidth}
         \centering
         \includegraphics[width=3.5cm,height=3.5cm]{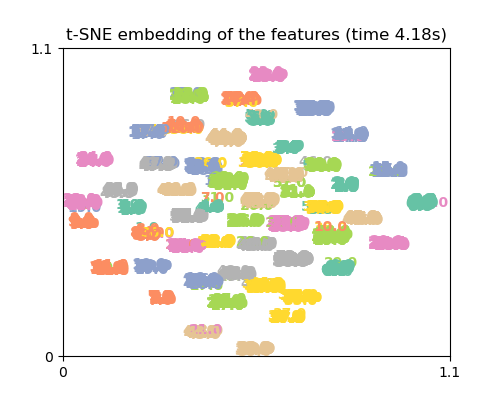}
         \caption{Low-Level Features \\(SDUgait)}
         \label{tsne:SDUgait1}
     \end{subfigure}
     \begin{subfigure}[b]{0.19\textwidth}
         \centering
         \includegraphics[width=3.5cm,height=3.5cm]{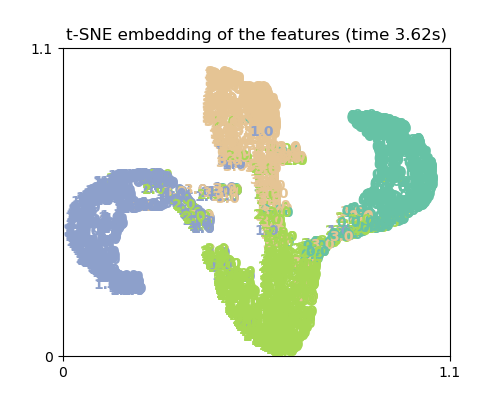}
         \caption{Low-Level Features \\(NDDs)}
         \label{tsne:NDDs1}
     \end{subfigure}
    \begin{subfigure}[b]{0.19\textwidth}
         \centering
         \includegraphics[width=3.5cm,height=3.5cm]{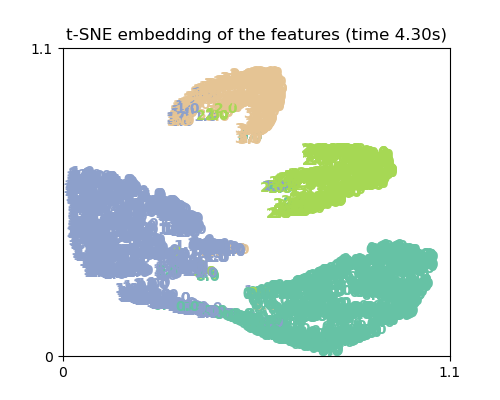}
         \caption{Low-Level Features \\(PD)}
         \label{tsne:PD1}
     \end{subfigure}
     \begin{subfigure}[b]{0.19\textwidth}
         \centering
         \includegraphics[width=3.5cm,height=3.5cm]{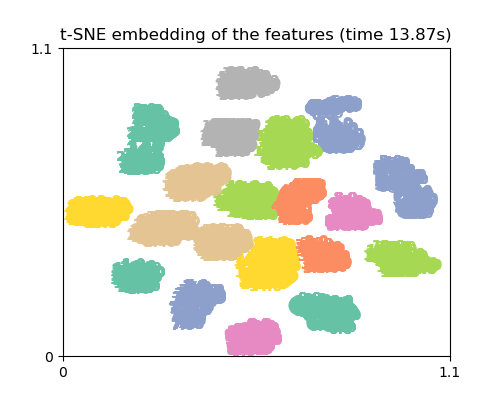}
         \caption{High-Level Features \\(UNITO)}
         \label{tsne:UNITO2}
     \end{subfigure}
     \begin{subfigure}[b]{0.19\textwidth}
         \centering
         \includegraphics[width=3.5cm,height=3.5cm]{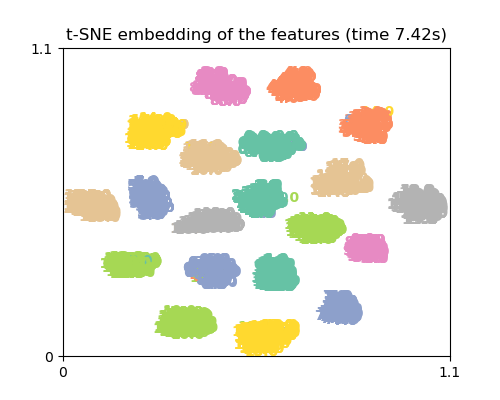}
         \caption{High-Level Features \\(CASIA-A)}
         \label{tsne:CASIA2}
     \end{subfigure}
    \begin{subfigure}[b]{0.19\textwidth}
         \centering
         \includegraphics[width=3.5cm,height=3.5cm]{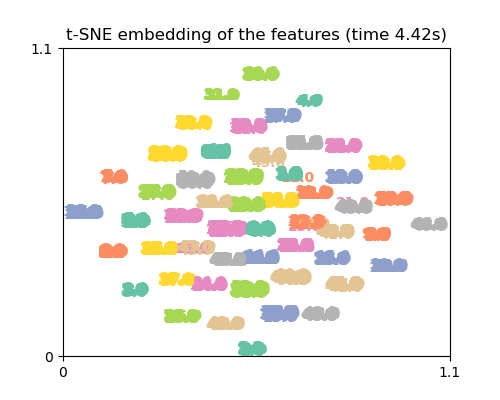}
         \caption{High-Level Features \\(SDUgait)}
         \label{tsne:SDUgait2}
     \end{subfigure}
     \begin{subfigure}[b]{0.19\textwidth}
         \centering
         \includegraphics[width=3.5cm,height=3.5cm]{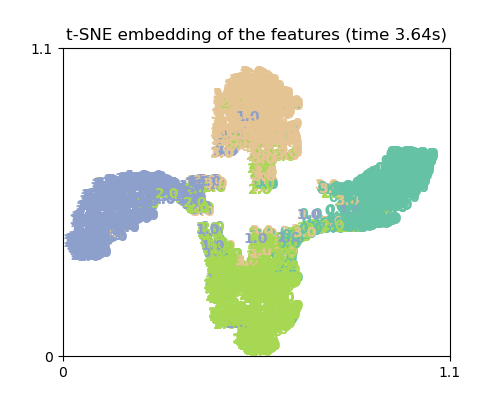}
         \caption{High-Level Features \\(NDDs)}
         \label{tsne:NDDs2}
     \end{subfigure}
    \begin{subfigure}[b]{0.19\textwidth}
         \centering
         \includegraphics[width=3.5cm,height=3.5cm]{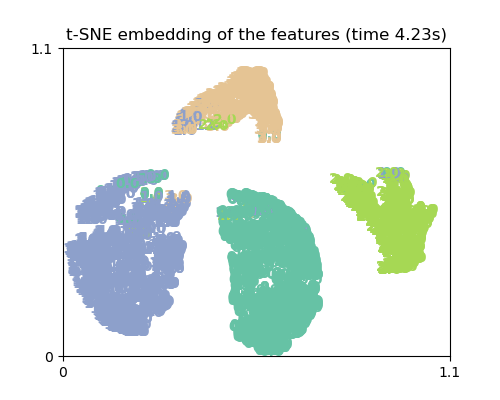}
         \caption{High-Level Features \\(PD)}
         \label{tsne:PD2}
     \end{subfigure}
\caption{Dimensionality reduction comparison between the extracted features and original data using t-SNE.}
\label{tsne}
\end{figure*}

We discuss the separability of the features extracted by ASTCapsNet, and take the outputs of $softmax~2$  and $softmax~4$ in the low- and high-level feature extractors as an example. All the features are processed by t-SNE (t-distributed stochastic neighbor embedding) for dimensionality reduction. As shown in Fig. \ref{tsne}, the first row represents the performance of the low-level extractor on the five datasets after dimensionality reduction, and the second row represents the performance of the high-level extractor after dimensionality reduction. We can clearly observe that the high-level features are more discriminative than the low-level features, and can represent the critical information of each class. The overlapping probability of low-level features is slightly larger than that of the high-level features. The high-level features have clear boundaries with the other classes after dimensionality reduction and have a high degree of aggregation with the samples in the same class, demonstrating the effectiveness of our proposed method.

\section{Discussion}

\begin{table*}[!htb]
\centering
\caption{Performance of different layers in ASTCapsNet.}
\scalebox{1}{
\begin{tabular}{lcccccc}
  \hline
  \hline
 Dataset&Softmax 1&Softmax 2&Softmax 3&Softmax 4&Total\\
  \hline
  UNITO&95.99\%&95.87\%&99.12\%&99.56\%&\textbf{\textit{99.65\%}}\\
  CASIA Gait&96.65\%&97.32\%&98.22\%&99.23\%&\textbf{\textit{99.69\%}}\\
  SDUgait&95.30\%&95.33\%&98.77\%&98.56\%&\textbf{\textit{99.13\%}}\\
  NDDs&93.86\%&93.64\%&94.98\%&95.33\%&\textbf{\textit{95.78\%}}\\
  PD&94.02\%&94.37\%&96.43\%&96.38\%&\textbf{\textit{97.31\%}}\\
\hline
\hline
\end{tabular}}
  \label{tb_split}
\end{table*}
As shown in the experiment, ASTCapsNet can effectively learn the human gait pattern including healthy people and NDDs patients. By experimenting on five gait datasets, the highest accuracy of our model are 99.65\%, 99.69\%, 99.13\%, 95.78\%, 97.31\% respectively, superior to the other gait recognition methods on the classification of these datasets. To evaluate the performance of each component in ASTCapsNet, we compare the classification results of each softmax layer with the output of the Bayesian model, shown in Table \ref{tb_split}. We can find that the performance of softmax 4 surpasses the other layers because this layer contains rich features calculated from the previous layers. Softmax 3 also performs well, reflecting the validity of the relationship layer. Furthermore, the model parameters have been found to affect the performance of classification, and the optimal setting is reported in Section IV.

In the field of gait recognition, there is still a lack of public datasets that have both large view variations and diversity of data. This factor affects the possibility to train a unified and reliable model for practical scenarios in medical diagnosis. For example, if the data of one modality is missing in the training, the final classification system may not be able to effectively handle this modality in testing.  Moreover, it is expensive to obtain the labeled gait sequences. In this section, we summarize several possible ways to improve the practicality and effectiveness of our proposed model as follows.

\textbf{A. Cross-view gait recognition.} In our ASTCapsNet, we only recognize the gait from different sources, without considering the relationship between multi-modal gait data. \cite{8374898} and \cite{8466612} have undertaken research and discussion on cross-view and multi-modal gait recognition. We will use canonical correlation analysis (CCA), kernel CCA (KCCA), and other data fusion methods to analyze the correlation between different sensors or views.

\textbf{B. Unsupervised or Semi-Supervised Learning.} The model discussed in this paper only refers to the encoder, which does not have the function of generating new samples and cannot deal with unlabeled gait data. How to utilize unlabeled samples for unsupervised or semi-supervised learning is still an open problem in this study. \cite{8374898} reports multi-task generative adversarial networks (MGANs) for learning view-specific feature representations, which provides a possible way to utilize unlabeled data in our model. We will explore to enhance our method by exploiting unlabeled information in future work.

\section{Conclusion}
In this paper, we have proposed an end-to-end learning model, ASTCapsNet, for gait recognition in multimodal gait sequences. The convolutional layer and a novel recurrent memory unit were introduced to model the dynamics and dependency in the spatio-temporal dimension for low-level feature extraction. Furthermore, a relationship layer was designed for high-level feature extraction, calculated between different capsules. Finally, a Bayesian model was also proposed for ASTCapsNet, with which our network can assess the quality of the features extracted from each layer using softmax. Our proposed method yields superior performance on the benchmark datasets for handling the gait recognition problem. This network will be extended to address the task of multi-view gait detection and recognition for intelligently streaming scalable video sequences, which requires analyzing gait changes in the video sequences and predicts the class of each person in real-time.






\footnotesize
\bibliographystyle{IEEEtran}
\bibliography{mybibfile}\ 

\end{document}